\documentclass[runningheads]{llncs}
\usepackage[T1]{fontenc}
\usepackage{graphicx}
\usepackage{hyperref}
\usepackage{color}

\usepackage{siunitx}
\usepackage{xcolor}
\usepackage{amsmath}
\usepackage{amssymb}
\usepackage{booktabs}
\usepackage{multirow}

\usepackage[misc]{ifsym}

\definecolor{myblue}{RGB}{0,176,240}
\definecolor{myred}{RGB}{192,0,0}
\definecolor{mygray}{RGB}{132,151,176}
\definecolor{mygold}{RGB}{255,192,0}

\begin{document}
\title{A denoised Mean Teacher for domain adaptive point cloud registration}
%
%
\author{Alexander Bigalke\textsuperscript{(\Letter)}\orcidID{0000-0001-7824-5735} \and
Mattias P. Heinrich\orcidID{0000-0002-7489-1972}}

%
\authorrunning{A. Bigalke et al.}
%
\institute{Institute of Medical Informatics, University of L\"ubeck, L\"ubeck, Germany \email{\{alexander.bigalke,mattias.heinrich\}@uni-luebeck.de}}
%
\maketitle              
\begin{abstract}
Point cloud-based medical registration promises increased computational efficiency, robustness to intensity shifts, and anonymity preservation but is limited by the inefficacy of unsupervised learning with similarity metrics.
Supervised training on synthetic deformations is an alternative but, in turn, suffers from the domain gap to the real domain.
In this work, we aim to tackle this gap through domain adaptation.
Self-training with the Mean Teacher is an established approach to this problem but is impaired by the inherent noise of the pseudo labels from the teacher.
As a remedy, we present a denoised teacher-student paradigm for point cloud registration, comprising two complementary denoising strategies.
First, we propose to filter pseudo labels based on the Chamfer distances of teacher and student registrations, thus preventing detrimental supervision by the teacher.
Second, we make the teacher dynamically synthesize novel training pairs with noise-free labels by warping its moving inputs with the predicted deformations.
Evaluation is performed for inhale-to-exhale registration of lung vessel trees on the public PVT dataset under two domain shifts.
Our method surpasses the baseline Mean Teacher by 13.5/62.8\%, consistently outperforms diverse competitors, and sets a new state-of-the-art accuracy (TRE=\SI{2.31}{mm}).
Code is available at \url{https://github.com/multimodallearning/denoised_mt_pcd_reg}.

\end{abstract}
\section{Introduction}
Recent deep learning-based registration methods have shown great potential in solving medical image registration problems \cite{fu2020deep,haskins2020deep}.
Most of these methods perform the registration based on the raw volumetric intensity images, e.g. \cite{balakrishnan2019voxelmorph,chen2022transmorph,de2019deep,mok2020large,zhao2019recursive}.
By contrast, only a few recent works \cite{hansen2021deep,shen2021accurate} operate on sparse, purely geometric point clouds extracted from the images, even though this representation promises multiple potential benefits, including computational efficiency, robustness against intensity shifts in the image domain, and anonymity preservation.
The latter, for instance, can facilitate public data access and federated learning, as exemplified by a recently released point cloud dataset of lung vessels \cite{shen2021accurate} whose underlying CT scans are not publicly accessible.
On the other hand, the sparsity of point clouds and the absence of intensity information make the registration problem more challenging.
In particular, unsupervised learning with similarity metrics -- as established for dense image registration \cite{chen2022transmorph,mok2020large} -- was shown ineffective for deformable point cloud registration \cite{shen2021accurate}, as confirmed by our experiments.
Since manual annotations for supervised learning are prohibitively costly, an alternative consists of training on synthetic deformations with known displacements \cite{shen2021accurate}, as known from dense registration \cite{eppenhof2018pulmonary,uzunova2017training}.
The inevitable domain gap between synthetic and real deformations, however, involves the risk of suboptimal performance on real data.
In this work, we aim to bridge this gap through domain adaptation (DA).

DA has widely been studied for classification and segmentation tasks \cite{guan2021domain}, with popular techniques ranging from adversarial feature \cite{ganin2015unsupervised,tzeng2017adversarial} or output \cite{tsai2018learning} alignment to self-supervised feature learning \cite{sun2019unsupervised}.
However, these methods are insufficient for the specific characteristics of the registration problem, involving a more complex output space and requiring the detection of local correspondences.
Instead, recent works adapted the Mean Teacher paradigm \cite{tarvainen2017mean}, previously established for domain adaptive classification \cite{french2018selfensembling} and segmentation \cite{perone2019unsupervised}, to the registration problem \cite{bigalke2022adapting,jin2022deformation,xu2022double}.
The basic idea is to supervise the learning student model with displacement fields (pseudo labels) provided by a teacher model, whose weights represent the exponential moving average of the student's weights.
A significant limitation of this method, however, is the inevitable noise in the pseudo labels, potentially misguiding the adaptation process.
Prior works addressed this problem by refining the pseudo labels \cite{jin2022deformation} or weighting them according to model uncertainty, estimated through Monte Carlo dropout \cite{xu2022double,yu2019uncertainty}.
However, even refined pseudo labels remain inaccurate, and the proposed refinement strategy \cite{jin2022deformation} assumes piecewise rigid motions of 3D objects and does not apply to complex deformations in medical applications.
And weighting pseudo labels according to teacher uncertainty \cite{xu2022double,yu2019uncertainty} does not explicitly consider the quality of the actual registrations, completely ignores the quality and certainty of the current student predictions, and can, therefore, not prevent detrimental supervision of the student through inferior teacher predictions.

\begin{figure}[t]
\centering
\includegraphics[width=\textwidth]{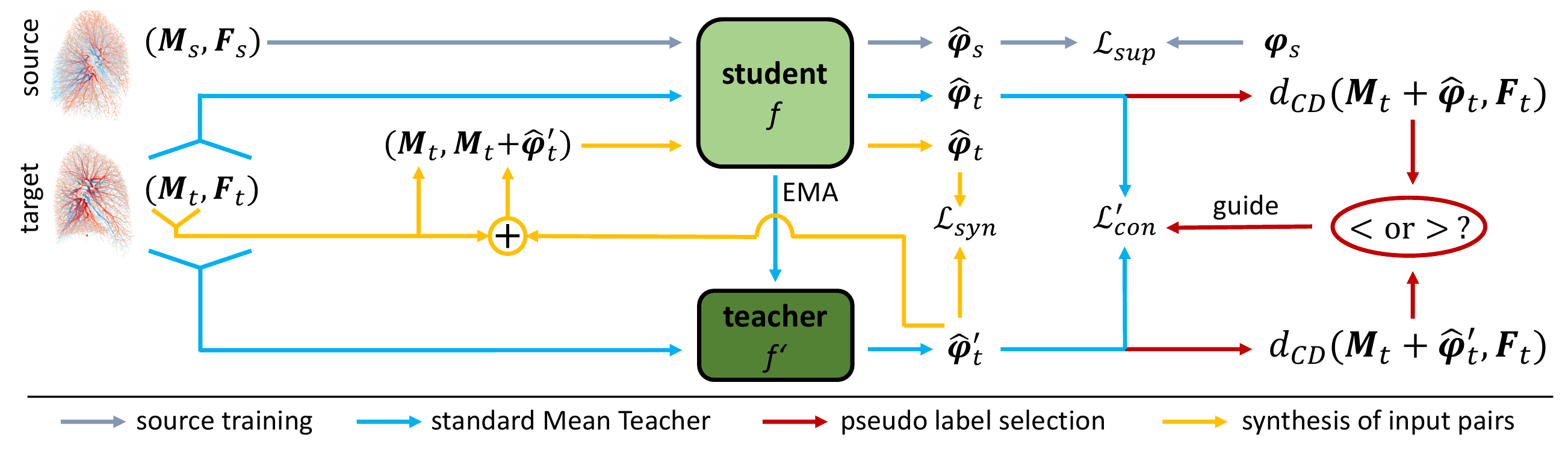}
\caption{Overview of our denoised Mean Teacher for domain adaptive registration. We overcome noisy supervision by the teacher with a novel pseudo label selection strategy and the synthesis of new training pairs with precisely known displacements.}
\label{fig:overview}
\end{figure}

\textbf{Contributions.} We introduce two complementary strategies to 
denoise the Mean Teacher for domain adaptive point cloud registration, addressing the above limitations (see Fig.~\ref{fig:overview}).
Both strategies are based on our understanding of an optimal student-teacher relationship.
First, if the student's solution to a problem is superior to that of the teacher, good teachers should not insist on their solution but accept the student's approach.
To implement this, inspired by a recent technique to filter pseudo labels for human pose estimation \cite{bigalke2022anatomy}, we propose to assess the quality of both the teacher and student registrations with the Chamfer distance and to provide only those registrations of the teacher as supervision to the student that are more accurate.
This approach differs from previous uncertainty-based methods \cite{xu2022double,yu2019uncertainty} in two decisive aspects:
1) It explicitly assesses the quality of final registrations, using a model-free and objective measure with little computational overhead compared to multiple forward passes in Monte Carlo dropout.
2) The selection process considers both teacher and student predictions and can thus prevent detrimental supervision by the teacher.
Our second strategy follows the intuition that good teachers should not pose problems to which they do not know the solution.
Instead, they should come up with novel tasks with precisely known solutions.
Consequently, we propose a completely novel teacher paradigm, where predicted deformations by the teacher are used to synthesize new training pairs for the student, consisting of the original moving inputs and their warps.
These input pairs come with precise noise-free displacement labels and significantly differ from static hand-crafted synthetic deformations \cite{shen2021accurate}.
1) The deformations are based on a real data pair that the teacher aims to align.
2) The deformations are dynamic and become more realistic as the teacher improves.
Finally, we unify both strategies in a joint framework for domain adaptive point cloud registration.
It is compatible with arbitrary geometric registration models, stable to train, and involves only a few hyper-parameters.
We experimentally evaluate the method for inhale-to-exhale registration of lung vessel point clouds on the public PVT dataset \cite{shen2021accurate}, demonstrating substantial improvements over diverse competing methods and state-of-the-art performance.

\section{Methods}
\subsection{Problem setup and standard Mean Teacher}
In point cloud registration, we are given fixed and moving point clouds $\boldsymbol{F}\in\mathbb{R}^{N_{\mathrm{F}}\times 3},\boldsymbol{M}\in\mathbb{R}^{N_{\mathrm{M}}\times 3}$ and aim to predict a displacement vector field $\boldsymbol{\varphi}\in\mathbb{R}^{N_{\mathrm{M}}\times 3}$ that spatially aligns $\boldsymbol{M}$ to $\boldsymbol{F}$ as $\boldsymbol{M}+\boldsymbol{\varphi}$.
We address the task in a domain adaptation setting with training data comprising a labeled source dataset $\mathcal{S}$ of triplets $(\boldsymbol{M}_s,\boldsymbol{F}_s,\boldsymbol{\varphi}_s)$ and a shifted unlabeled target dataset $\mathcal{T}$ of tuples $(\boldsymbol{M}_t,\boldsymbol{F}_t)$.
While the formulation of our method is agnostic to the specific domain shift between $\mathcal{S}$ and $\mathcal{T}$, in this work, we generate the source samples on the fly as random synthetic deformations of the target clouds using a fixed hand-crafted deformation function $def:\mathbb{R}^{N\times 3}\rightarrow\mathbb{R}^{N\times 3}$, i.e. source triplets are given as $(def(\boldsymbol{F}_t),\boldsymbol{F}_t,\boldsymbol{F}_t-def(\boldsymbol{F}_t))$ or $(def(\boldsymbol{M}_t),\boldsymbol{M}_t,\boldsymbol{M}_t-def(\boldsymbol{M}_t))$.
Note that $def$ preserves point correspondences enabling ground truth computation through point-wise subtraction.
Given the training data, we aim to learn a function $f$ that predicts deformation vector fields as $\hat{\boldsymbol{\varphi}}=f(\boldsymbol{M},\boldsymbol{F})$ with optimal performance in the target domain.

\textbf{Baseline Mean Teacher.}
To solve the problem, the standard Mean Teacher framework \cite{bigalke2022adapting,french2018selfensembling,tarvainen2017mean} employs two identical networks, denoted as the student $f$ and teacher $f'$, with parameters $\boldsymbol{\theta}$ and $\boldsymbol{\theta}'$.
While the student's weights $\boldsymbol{\theta}$ are optimized through gradient descent, the teacher's weights correspond to the exponential moving average (EMA) of the student and are updated as $\boldsymbol{\theta}'_i=\alpha\boldsymbol{\theta}'_{i-1}+(1-\alpha)\boldsymbol{\theta}_i$ at iteration $i$ with momentum $\alpha$. 
Meanwhile, the student is trained by minimizing
\begin{equation}
\label{eq:baseline_loss}
    \mathcal{L}(\boldsymbol{\theta})=\lambda_1\underbrace{\|f(\boldsymbol{M}_s,\boldsymbol{F}_s)-\boldsymbol{\varphi}_s\|_2^2}_{\mathcal{L}_{\mathrm{sup}}}+\lambda_2\underbrace{\|f(\boldsymbol{M}_t,\boldsymbol{F}_t)-f'(\boldsymbol{M}_t,\boldsymbol{F}_t)\|_2^2}_{\mathcal{L}_{\mathrm{con}}}
\end{equation}
consisting of the supervised loss $\mathcal{L}_{\mathrm{sup}}$ on source data and the consistency loss $\mathcal{L}_{\mathrm{con}}$ on target data, weighted by $\lambda_1$ and $\lambda_2$.
$\mathcal{L}_{\mathrm{con}}$ guides the learning of the student in the target domain with pseudo-supervision from the teacher, which, as a temporal ensemble, is expected to be superior to the student.
Nonetheless, predictions by the teacher can still be noisy and inaccurate, limiting the efficacy of the adaptation process.

\subsection{Chamfer distance-based filtering of pseudo labels}
In a worst-case scenario, the student might predict an accurate displacement field $\hat{\boldsymbol{\varphi}}_t$, which is strongly penalized by the consistency loss due to an inaccurate teacher prediction $\hat{\boldsymbol{\varphi}}'_t$.
To prevent such detrimental supervision, we aim to select only those teacher predictions for supervision that are superior to the corresponding student predictions, which, however, is complicated by the absence of ground truth.
We, therefore, propose to assess the quality of student and teacher registrations by measuring the similarity/distance between fixed and warped moving clouds, with higher similarities/lower distances indicating more accurate registrations.
Among existing similarity measures, we opt for the symmetric Chamfer distance \cite{wu2020pointpwc}, which computes the distance between two point clouds $\boldsymbol{X},\boldsymbol{Y}$ as
\begin{equation}
    d_{\mathrm{CD}}(\boldsymbol{X},\boldsymbol{Y})=\sum_{\boldsymbol{x}\in\boldsymbol{X}}\mathrm{min}_{\boldsymbol{y}\in\boldsymbol{Y}}\|\boldsymbol{x}-\boldsymbol{y}\|_2^2+\sum_{\boldsymbol{y}\in\boldsymbol{Y}}\mathrm{min}_{\boldsymbol{x}\in\boldsymbol{X}}\|\boldsymbol{x}-\boldsymbol{y}\|_2^2
\end{equation}
While we experimentally found the Chamfer distance insufficient as a direct loss function -- presumably due to sparse differentiability and susceptibility to local minima, we still observed a strong correlation between Chamfer distance and actual registration error, making it a suitable choice for our purposes.
We also explored other measures (Laplacian curvature \cite{wu2020pointpwc}, Gaussian MMD \cite{feydy2020geometric}), which proved slightly inferior (Supp., Tab.~\ref{tab:results_supp}).
Formally, we thus measure the quality of the student prediction $\hat{\boldsymbol{\varphi}}_t=f(\boldsymbol{M}_t,\boldsymbol{F}_t)$ as $d_{\mathrm{CD}}(\boldsymbol{M}_t+\hat{\boldsymbol{\varphi}}_t,\boldsymbol{F}_t)$ and analogously for the teacher prediction $\hat{\boldsymbol{\varphi}}'_t$.
We then define our indicator function
\begin{equation}
    I(\hat{\boldsymbol{\varphi}}_t,\hat{\boldsymbol{\varphi}}'_t)=
    \begin{cases}
    1\qquad d_{\mathrm{CD}}(\boldsymbol{M}_t+\hat{\boldsymbol{\varphi}}'_t,\boldsymbol{F}_t)\, <\, d_{\mathrm{CD}}(\boldsymbol{M}_t+\hat{\boldsymbol{\varphi}}_t,\boldsymbol{F}_t)\\
    0\qquad \mathrm{else}
    \end{cases}
\end{equation}
and reformulate the consistency loss in Eq.~\ref{eq:baseline_loss} as
\begin{equation}
    \mathcal{L}'_{\mathrm{con}}=I(\hat{\boldsymbol{\varphi}}_t,\hat{\boldsymbol{\varphi}}'_t)\cdot\|f(\boldsymbol{M}_t,\boldsymbol{F}_t)-f'(\boldsymbol{M}_t,\boldsymbol{F}_t)\|_2^2
\end{equation}

\subsection{Synthesizing inputs with noise-free supervision}
\label{sec:generative_mt}
While the above filtering strategy mitigates detrimental supervision, the selected pseudo labels are still inaccurate.
Therefore, we complement the strategy with a novel teacher paradigm, where the teacher dynamically synthesizes new training pairs with precisely known displacements for supervision.
Specifically, given a teacher prediction $\hat{\boldsymbol{\varphi}}'_t=f'(\boldsymbol{M}_t,\boldsymbol{F}_t)$, we do not only use it to supervise the student on the same input pair but also generate a new input sample $(\boldsymbol{M}_t,\boldsymbol{M}_t+\hat{\boldsymbol{\varphi}}'_t)$ by warping $\boldsymbol{M}_t$ with $\hat{\boldsymbol{\varphi}}'_t$.
The underlying displacement field is naturally precisely known, enabling noise-free training of the student by minimizing
\begin{equation}
    \mathcal{L}_\mathrm{syn}=\|f(\boldsymbol{M}_t,\boldsymbol{M}_t+\hat{\boldsymbol{\varphi}}'_t)-\hat{\boldsymbol{\varphi}}'_t\|^2_2
\end{equation}
To our knowledge, there is no prior work with a similarly ``generative'' teacher model.
Altogether, we train the student network by minimizing the loss
\begin{equation}
\label{eq:total_loss}
    \mathcal{L}(\boldsymbol{\theta})=\lambda_1\mathcal{L}_{\mathrm{sup}}+\lambda_2\mathcal{L}'_{\mathrm{con}}+\lambda_3\mathcal{L}_{\mathrm{syn}}
\end{equation}

\textbf{Technical details.}
The synthesized input pairs $(\boldsymbol{M}_t,\boldsymbol{M}_t+\hat{\boldsymbol{\varphi}}'_t)$ exhibit exact point correspondence, i.e., for each point in $\boldsymbol{M}_t$ exists a corresponding point in $\boldsymbol{M}_t+\hat{\boldsymbol{\varphi}}'_t$.
That is usually not the case for real data pairs and thus introduces another domain shift, which prevented proper convergence in our initial experiments.
To overcome the problem, we exploit that the original point clouds in a dataset, denoted as $\boldsymbol{M}_{t*}$, usually comprise more points than the subsampled clouds $\boldsymbol{M}_t$ that are fed to the network.
Given predicted displacements $\hat{\boldsymbol{\varphi}}'_t$ for $\boldsymbol{M}_t$, we interpolate the displacement vectors to $\boldsymbol{M}_{t*}$ with an isotropic Gaussian kernel, yielding $\hat{\boldsymbol{\varphi}}'_{t*}$.
The final input pair is then obtained by sampling disjoint point subsets from $(\boldsymbol{M}_{t*},\boldsymbol{M}_{t*}+\hat{\boldsymbol{\varphi}}'_{t*})$, excluding one-to-one correspondences.

\section{Experiments}
\subsection{Experimental setup}
\subsubsection{Datasets.}
We evaluate our method for inhale-to-exhale registration of lung vessel point clouds on the public PVT dataset \cite{shen2021accurate} (\url{https://github.com/uncbiag/robot}, License: CC BY-NC-SA 3.0).
The dataset comprises 1,010 such data pairs, which were extracted from lung CT scans as part of the IRB-approved COPDGene study (NCT00608764).
Ten of these scan pairs are cases from the Dirlab-COPDGene dataset \cite{castillo2013reference} and thus annotated with 300 landmark correspondences.
We use these cases as the test set and split the remaining unlabeled pairs into 800 cases for training and 200 for validation (on synthetic deformations only).
The original point clouds in the dataset have a very high resolution ($\sim$100k points), making the processing with deep networks computationally costly.
Therefore, we extract distinctive keypoints by local density estimation followed by non-maximum suppression.
We extract two sets of such keypoints for each cloud: one with the $\sim$8k most distinctive points for inference, and another with $\sim$16k points, from which we randomly sample subsets during training for increased variability (see Sec.~\ref{sec:generative_mt}, technical details).
Finally, we pre-align each pair by matching the mean and standard deviation of the coordinates.

\subsubsection{Implementation details.}
The registration network $f$ is implemented as the default 4-scale architecture of PointPWC-Net \cite{wu2020pointpwc}, operating on 8192 points per cloud.
Following \cite{wu2020pointpwc}, we implement $\mathcal{L}_{\mathrm{sup}}$, $\mathcal{L}'_{\mathrm{con}}$, and $\mathcal{L}_{\mathrm{syn}}$ as multi-scale losses.
Optimization is performed with the Adam optimizer.
We first pre-train the network on source data (batch size 4) for 160 epochs and subsequently minimize the joint loss (Eq.~\ref{eq:total_loss}) for 140 epochs, both with a constant learning rate of 0.001, which requires up to \SI{11}{GB} and 13/\SI{23}{h} on an RTX2080.
For joint optimization, we use mixed batches of 4 source and 4 target samples, set $\lambda_1=\lambda_2=\lambda_3=10$, and the EMA-parameter to $\alpha=0.996$.
While the original PVT data pairs represent the target domain in all experiments, we consider two variants of the function $def$ to synthesize source data pairs: a realistic task-specific 2-scale random field similar to \cite{shen2021accurate} and a simple rigid transformation.
This enables us to evaluate our method under two differently severe domain shifts.
Since real validation data are unavailable, hyper-parameters of all compared methods were tuned in a synthetic adaptation scenario, with the rigid deformations in the source and the 2-scale random field deformations in the target domain.
For further implementation details, we refer to our public code.

\subsubsection{Comparison methods.}
\label{sec:comparison_methods}
1) The source-only model is exclusively trained on source data without DA.
2) We adopt the standard Mean Teacher \cite{bigalke2022adapting}.
3) An uncertainty-aware Mean Teacher (UA-MT), similar to \cite{xu2022double,yu2019uncertainty}.
4) As proposed in \cite{wu2020pointpwc}, we performed purely unsupervised training on target data with a Chamfer loss. 
However, consistent with the findings in \cite{shen2021accurate}, this approach could not converge for complex geometric lung structures.
Instead, we use the Chamfer loss on target data as an additional loss to complement supervised source training.
5) We guide the learning on target data with the cycle-consistency method from \cite{mittal2020just}.
6) As a classical algorithm, we adapt sLBP \cite{hansen2021deep}.
7) We collect the results of two current SOTA methods, S-Robot and D-Robot, from \cite{shen2021accurate}, which combine deep networks (Point U-Net, PointPWC-Net), trained on synthetic deformations, with optimal transport modules.
Note, however, that the experimental setup in \cite{shen2021accurate} slightly differs from our setting in terms of more input points (60k vs.~8k) and additional input features (vessel radii), thus accessing more information.

\subsubsection{Metrics.}
We interpolate the predicted displacements from the moving input cloud to the annotated moving landmarks with an isotropic Gaussian kernel ($\sigma=\SI{5}{mm}$) and measure the target registration error (TRE) with respect to the fixed landmarks.
To assess the smoothness of the predictions, we interpolate the sparse displacement fields to the underlying image grid and measure the standard deviation of the logarithm of the Jacobian determinant (SDlogJ).

\subsection{Results}
\begin{figure}[t]
\centering
\includegraphics[width=\textwidth]{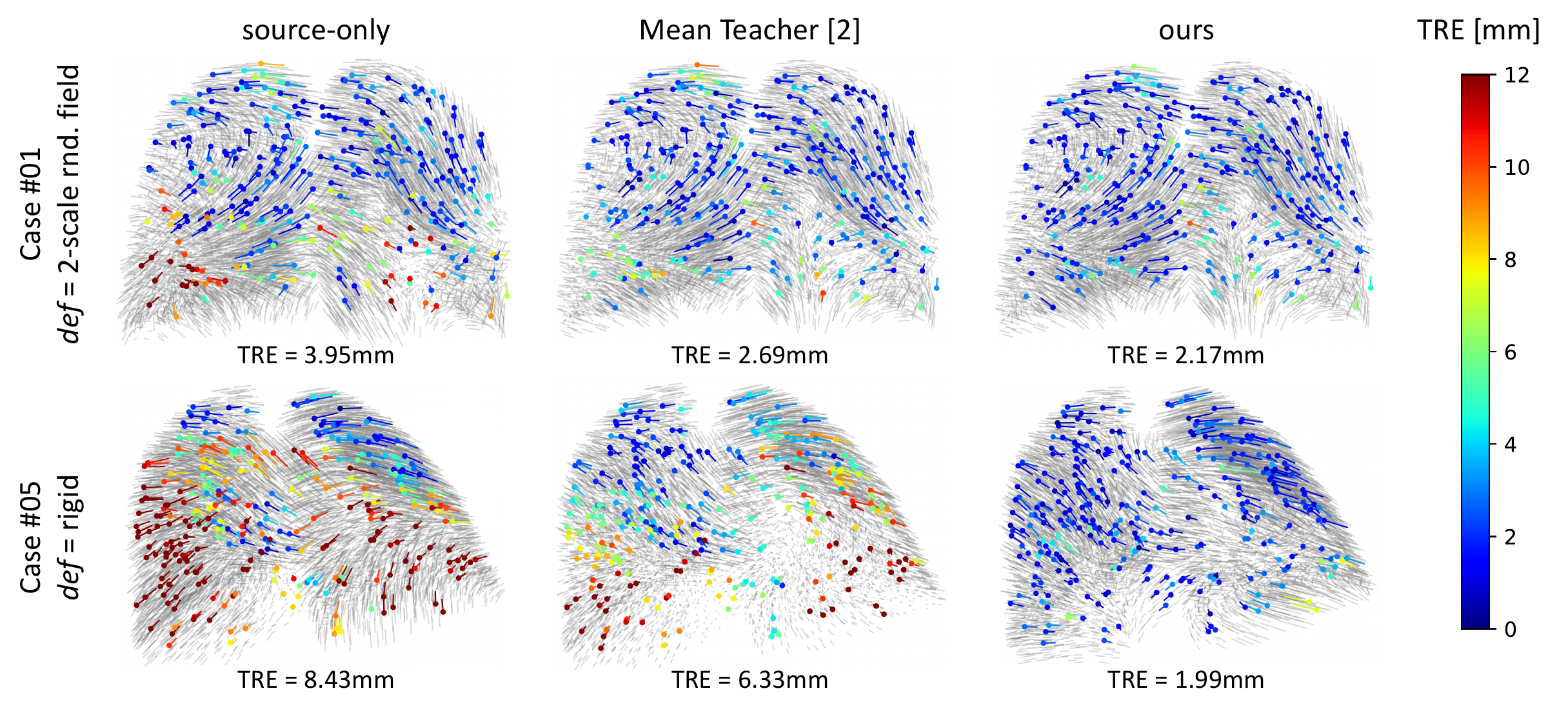}
\caption{
Qualitative results on two sample cases of the PVT dataset.
Predicted displacement fields are shown in gray.
Colored dots and lines represent moving landmarks and their interpolated flow, with colors encoding the TRE (clamped to \SI{12}{mm}).
}
\label{fig:qual_results}
\end{figure}

\begin{figure}[t]
\centering
\includegraphics[width=\textwidth]{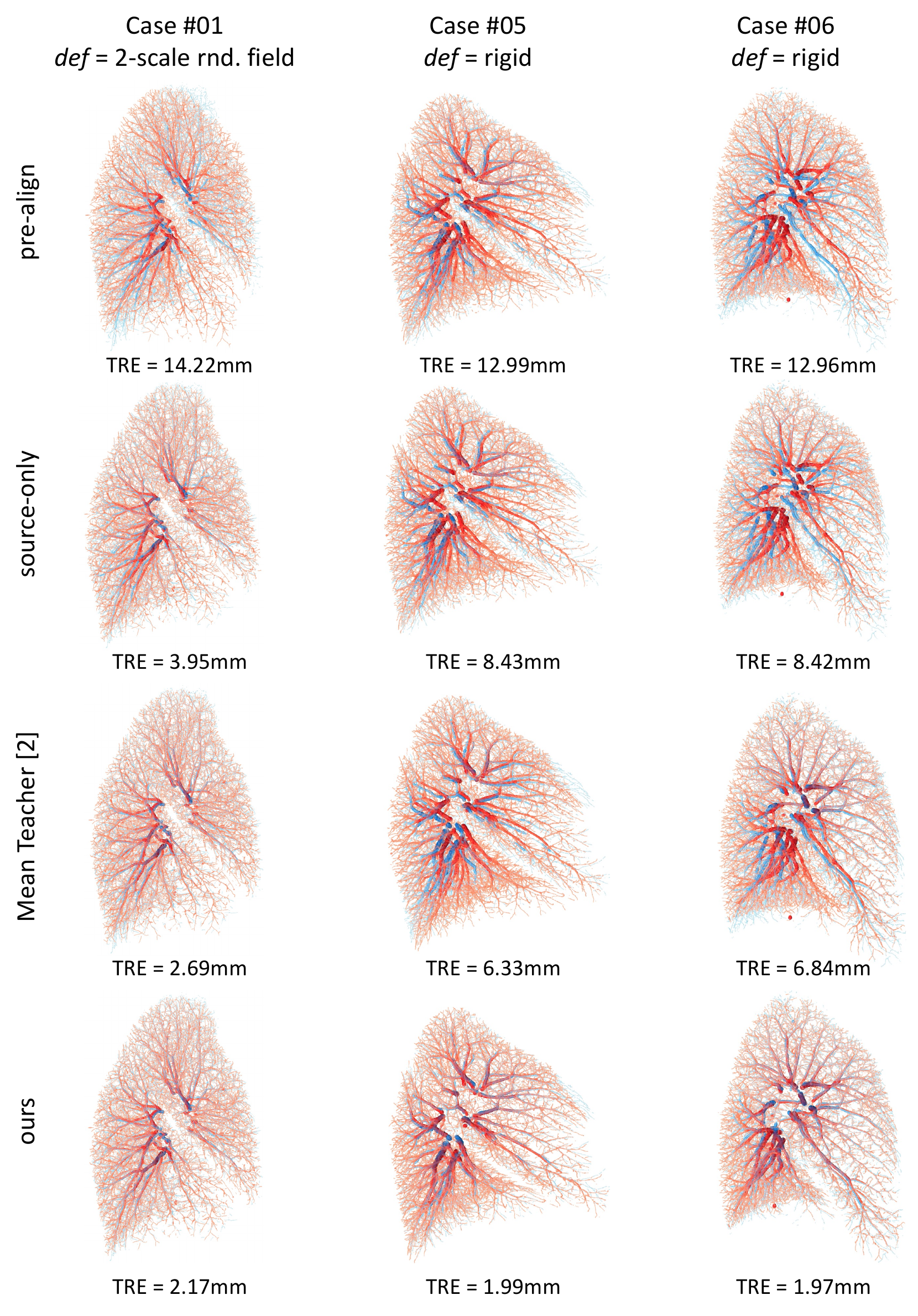}
\caption{Qualitative results on three cases of the PVT dataset.
	We show overlays of the original high-resolution point clouds of the fixed exhale and warped inhale structures in blue and red, respectively.
    To this end, we interpolated predicted displacement fields to the high-resolution inhale clouds with an isotropic Gaussian kernel.
	From top to bottom, each column shows the pre-aligned point clouds and the registrations by the source-only model, the standard Mean Teacher, and our method.}
\label{fig:qual_results_supp}
\end{figure}

\begin{table}[t]
\setlength{\tabcolsep}{4pt}
\caption{Quantitative results on the PVT dataset, reported as mean TRE and 25/75\% percentiles in mm and SDlogJ. $^{\dag}$ indicates a deviating experimental setup (Sec.~\ref{sec:comparison_methods}).}
\label{tab:results}
\centering
\begin{tabular}{lcccccccc}
\toprule
\multirow{2}{*}{Method} & \multicolumn{4}{c}{$def$ = 2-scale rnd.~field} & \multicolumn{4}{c}{$def$ = rigid}\\
    & TRE & 25\% & 75\% & SDlogJ & TRE & 25\% & 75\% & SDlogJ\\
\midrule
initial & 23.32 & 13.22 & 31.61 & - & 23.32 & 13.22 & 31.61 & -\\
pre-align & 12.83 & 8.25 & 16.68 & - & 12.83 & 8.25 & 16.68 & -\\
\midrule
sLBP \cite{hansen2021deep} & 3.62 & 1.24 & 3.29 & 0.038 & 3.62 & 1.24 & 3.29 & 0.038\\
S-Robot$^{\dag}$ \cite{shen2021accurate} & 5.48 & 2.86 & 7.14 & N/A & N/A & N/A & N/A & N/A\\
D-Robot$^{\dag}$ \cite{shen2021accurate} & 2.86 & 1.25 & 3.11 & N/A & N/A & N/A & N/A & N/A\\
\midrule
source-only & 4.50 & 1.62 & 5.49 & 0.034 & 7.62 & 3.12 & 11.15 & 0.019\\
Chamfer loss \cite{wu2020pointpwc} & 3.96 & 1.47 & 4.43 & 0.036 & 4.18 & 1.54 & 5.27 & 0.043\\
cycle-consistency \cite{mittal2020just} & 3.93 & 1.48 & 4.36 & 0.035 & 6.47 & 2.43 & 9.08 & 0.029\\ 
Mean Teacher \cite{bigalke2022adapting} & 2.67 & 1.33 & 3.12 & $\boldsymbol{0.028}$ & 6.40 & 2.42 & 9.50 & $\boldsymbol{0.013}$\\
UA-MT \cite{xu2022double} & 2.58 & 1.28 & 3.04 & 0.029 & 5.71 & 1.83 & 8.77 & 0.015\\
\midrule
ours w/o $\mathcal{L}_{\mathrm{syn}}$ & 2.49 & 1.23 & 2.88 & 0.030 & 2.57 & 1.22 & 2.93 & 0.027\\
ours w/o $\mathcal{L}'_{\mathrm{con}}$ & 2.96 & 1.27 & 3.29 & 0.035 & 3.00 & 1.21 & 3.39 & 0.034\\
ours & $\boldsymbol{2.31}$ & $\boldsymbol{1.16}$ & $\boldsymbol{2.66}$ & 0.034 & $\boldsymbol{2.38}$ & $\boldsymbol{1.12}$ & $\boldsymbol{2.66}$ & 0.033\\
\bottomrule
\end{tabular}
\end{table}
Quantitative results are shown in Tab.~\ref{tab:results} and reveal the following insights:
1) The source-only model benefits from realistic synthetic deformations in the source domain, yielding a 40.9\% lower TRE.
2) The standard Mean Teacher proves effective under the weaker domain shift (\mbox{$-40.7$\%} TRE compared to source-only) but only achieves a slight improvement of 16.0\% in the more challenging scenario, where pseudo labels by the teacher are naturally noisier, in turn limiting the efficacy of the adaptation process.
3) Our proposed strategy to filter pseudo labels (ours w/o $\mathcal{L}_{\mathrm{syn}}$) improves the standard teacher and its uncertainty-aware extension, particularly notable under the more severe domain shift (\mbox{$-59.8$}/\mbox{$-55.0$\%} TRE).
4) Synthesizing novel data pairs with the teacher (ours w/o $\mathcal{L}'_{\mathrm{con}}$) alone is slightly inferior to the standard teacher for realistic deformations in the source domain but substantially superior for simple rigid transformations.
5) Combining our two strategies yields further considerable improvements to TREs of 2.31 and \SI{2.38}{mm}, demonstrating their complementarity.
Thus, our method improves the standard Mean Teacher by 13.5/62.8\%, outperforms all competitors by statistically significant margins ($p<0.001$ in a Wilcoxon signed-rank test), and sets a new state-of-the-art accuracy.
Remarkably, our method achieves almost the same accuracy for simple rigid transformations in the source domain as for complex, realistic deformations.
Thus, it eliminates the need for designing task-specific deformation models, which requires strong domain knowledge.
Qualitative results are presented in Fig.~\ref{fig:qual_results} and Supp., Fig.~\ref{fig:qual_results_supp}, demonstrating accurate and smooth deformation fields by our method, as confirmed by the SDlogJ in Tab.~\ref{tab:results}, which takes small values for all methods.

\section{Conclusion}
Our work addressed domain adaptive point cloud registration to bridge the gap between synthetic source and real target deformations.
Starting from the established Mean Teacher paradigm, we presented two novel strategies to tackle the noise of pseudo labels from the teacher model, which is a persistent, significant limitation of the method.
Specifically, we 1) proposed to prevent detrimental supervision through the teacher by filtering pseudo labels according to Chamfer distances of student and teacher registrations and 2) introduced a novel teacher-student paradigm, where the teacher synthesizes novel training data pairs with perfect noise-free displacement labels.
Our experiments for lung vessel registration on the PVT dataset demonstrated the efficacy of our method under two scenarios, outperforming the standard Mean Teacher by up to 62.8\% and setting a new state-of-the-art accuracy (TRE=\SI{2.31}{mm}).
As such, our method even favorably compares to popular image-based deep learning methods (VoxelMorph \cite{balakrishnan2019voxelmorph} and LapIRN \cite{mok2020large}, e.g., achieve TREs of 7.98 and \SI{4.99}{mm} on the original DIR-Lab CT images) but lags behind conventional image-based optimization methods \cite{ruhaak2017estimation} with \SI{0.83}{mm} TRE.
But while the latter require run times of several minutes to process the dense intensity scans with 30M+ voxels, our method processes sparse, purely geometric point clouds with 8k points only, enabling anonymity-preservation and extremely fast inference within \SI{0.2}{s}.
In this light, we see two significant potential impacts of our work:
First, our method could generally advance purely geometric keypoint-based medical registration, previously limited by the inefficacy of unsupervised learning with similarity metrics.
In particular, medical point cloud registration, currently primarily focusing on lung anatomies, still needs to be investigated for other anatomical structures (abdomen, brain) in future work, which might benefit from our generic approach.
Second, our method is conceptionally transferable to dense image registration (e.g., intensity-based similarity metrics \cite{heinrich2012mind,de2020mutual} can replace the Chamfer distance).
In this context, it appears of great interest to revisit learning from synthetic deformations \cite{eppenhof2018pulmonary} within a DA setting or to combine our method with unsupervised learning under metric supervision.

\paragraph{Acknowledgement.} We gratefully acknowledge the financial support by the \linebreak Federal Ministry for Economic Affairs and Climate Action of Germany \linebreak (FKZ: 01MK20012B) and by the Federal Ministry for Education and Research of Germany (FKZ: 01KL2008).

%
%
\bibliographystyle{splncs04}
\bibliography{paper1168-bibliography}

\newpage
\section*{Supplementary Material}
\begin{table}[h]
\setlength{\tabcolsep}{6pt}
\caption{Quantitative results for different similarity measures to filter pseudo labels.
As metrics, we report the mean TRE and 25\%/75\% percentiles in mm.
We performed the experiment for rigid deformations in the source domain ($def$ = rigid).}
\label{tab:results_supp}
\centering
\begin{tabular}{lccc}
\toprule
Similarity measure & TRE & 25\% & 75\%\\
\midrule
none (standard Mean Teacher) & 6.40 & 2.42 & 9.50\\
Gaussian MMD [7] & 2.72 & 1.28 & 3.19 \\
Laplacian curvature [27] & 2.86 & 1.34 & 3.43\\
Chamfer distance (ours w/o $\mathcal{L}_{\mathrm{syn}}$) & $\boldsymbol{2.57}$ & $\boldsymbol{1.22}$ & $\boldsymbol{2.93}$\\
\bottomrule
\end{tabular}
\end{table}

\end{document}